\documentclass[10pt, a4paper]{article}
\usepackage{hybridreport}

\usepackage{graphicx}
\graphicspath{{figs/}}
\usepackage[square,authoryear]{natbib}

\newcommand{\task}[1]{\textit{#1}}
\newcommand{\pyrph}[1]{\task{Pyramid} phase~#1}
\DeclareTextFontCommand{\textbf}{\bfseries}

\reporttype{Technical Report}
\title{RoboPace: Contact-Aware Time-Optimal Retiming for Action-Chunk Policies}

\author{
  \mbox{Mimo Shirasaka\textsuperscript{1,2}}\quad
  \mbox{Takehiko Ohkawa\textsuperscript{1}}\quad
  \mbox{Takuya Okubo\textsuperscript{1}}\quad
  \mbox{Nicola Scianca\textsuperscript{1}}\quad
  \mbox{Tatsuya Matsushima\textsuperscript{1,2}}\quad
  \mbox{Kei Ota\textsuperscript{1}}
}
\institution{
  \textsuperscript{1}AI Robot Association (AIRoA)\quad
  \textsuperscript{2}The University of Tokyo
}
\date{October 2026}
\logo{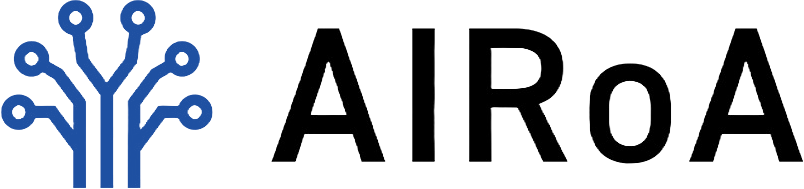}
\correspondence{%
  \href{mailto:shirasaka.mimo@airoa.org}{\texttt{shirasaka.mimo@airoa.org}}\\
  Project page: \href{https://research.airoa.io/publications/robopace/}{\texttt{https://research.airoa.io/publications/robopace/}}%
}
\abstracttext{Robot manipulation data collection has been shifting from teleoperation toward robot-free demonstrations, through interfaces such as the Universal Manipulation Interface (UMI) or directly from human hands. Vision-Language-Action (VLA) policies trained on such data inherit the demonstrator's timing. Yet human timing does not directly transfer to robots: compliant hands tolerate fast contact, whereas robots may overshoot due to actuator and tracking limitations; conversely, robots can move faster in free space. This motivates a unified approach that reconciles execution speed with contact safety.
We present RoboPace, an online retiming layer that preserves the policy's geometric path while adapting its timing, respecting the target robot's kinematic and dynamic constraints. It adapts execution speed based on predicted contact, jointly accounting for contact-dependent speed limits and the robot's motion constraints. The method requires no policy retraining and operates in real time. Across three contact-rich tasks on a dual-arm robot, faster uniform execution and physical-limit-only retiming largely fail. RoboPace instead achieves higher overall success than slow uniform execution while completing four of five commands in approximately half the time, retaining the reliability of slow execution without its time cost.
}

\begin{document}
\maketitle

\begin{figure}[H]
  \centering
  \includegraphics[width=\linewidth]{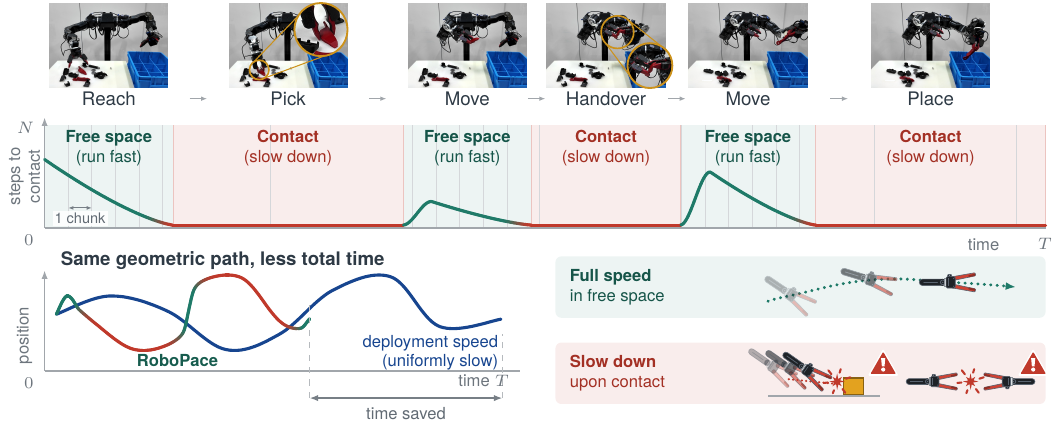}
  \caption{\textbf{Contact-Aware Retiming of Action Chunks:} RoboPace preserves the
  geometric path a policy predicts and adapts only its timing while respecting the
  robot's kinematic and dynamic constraints: predicted contact imposes a speed
  limit on the hands at contact-critical waypoints (red), while free-space motion
  (green) runs as fast as those constraints allow. On a dual-arm robot, this
  achieves higher overall success than uniformly slow execution while completing
  four of five commands in about half its time.}
  \label{fig:teaser}
\end{figure}

\section{Introduction}
\label{sec:intro}
Robot manipulation is increasingly benefiting from demonstrations collected
across embodiments. Teleoperated demonstrations on low-cost bimanual platforms
\cite{zhao2023aloha} have been complemented by handheld interfaces that
allow people to record demonstrations without a robot in the loop
\cite{chi2024umi, ohkawa2026yubi}, making data collection more
scalable. These interfaces record end-effector trajectories, from which policies
can learn to predict sequences of actions (action chunks) \cite{intelligence2025pi05} that are realized on a target robot. This flexibility introduces an \emph{embodiment gap} spanning morphology,
kinematics, and dynamics \cite{ohkawa2026yubi}. Successful deployment depends not only on whether the robot can reproduce a
demonstrated motion, but also on how quickly and reliably it can execute it.

The timing encoded in a demonstration is inherited from the demonstrator but does not necessarily transfer to the target robot. Near contact, the same motion may be too fast for a robot with different compliance, actuator limits, and tracking dynamics, leading to overshoot or unintended impact. In free space, the mismatch runs the other way: the demonstrator's pace is no bound on the robot, which could move faster. A single global speed factor cannot resolve this trade-off. Slowing the whole trajectory makes contact more reliable but gives
up the speed available elsewhere, while speeding it up increases the risk of failure near contact.

Resolving this trade-off requires execution speed to vary along the predicted motion according to predicted contact, while respecting the target robot's kinematic and dynamic constraints. Prior work adapts execution timing at training time or
inference time. Training-time methods modify demonstration timing through acceleration or downsampling based on entropy or semantic criteria \cite{guo2025demospeedup, kim2026espada}, or explicitly represent execution tempo in the policy \cite{ijspeert2013dmp, nam2026speedaug}. Such approaches can improve execution efficiency, but adapting to a target robot with different dynamic limits may require modifying the training data or retraining the policy. Inference-time methods instead adapt execution while leaving the policy unchanged. Time-optimal path parameterization (TOPP) \cite{pham2018toppra} has been applied to chunked policies \cite{kim2026race} to retime predicted paths under the robot's kinematic and dynamic constraints, but has no notion of contact. Criticality-based speed adaptation \cite{ranawaka2025sail} instead switches between preset global rates based on motion complexity or gripper events. Neither line of work jointly derives a contact-dependent speed constraint along the predicted motion and incorporates it into retiming under the target robot's kinematic and dynamic constraints.

We address this gap with RoboPace, an inference-time retiming method that imposes a position-dependent velocity ceiling that limits speed near predicted contact and incorporates it with the robot's kinematic and dynamic constraints in a single time-optimal path parameterization. RoboPace predicts contact using two complementary cues: geometric proximity and wrist-camera prediction. Forward kinematics places the robot's hand meshes
along the predicted path and measures proximity to the other hand and the
table, limiting speed when either is approached. The wrist-camera model
predicts the remaining steps until hand-to-object contact. Either cue can
tighten the velocity ceiling, with smooth transitions back to full speed away
from contact. RoboPace retimes newly predicted action chunks online during execution, requires no policy
retraining, and sits between trajectory generation and low-level tracking
without modifying either the policy or the tracking controller.

We evaluate RoboPace in real-world experiments on a dual-arm robot across five
commands spanning three contact-rich tasks, \task{Cup}, \task{Socks}, and
\task{Pyramid}. We compare against the $30$\,Hz rollout rate used in training
(the demo speed hereafter), uniform speed scaling, and physical-limit-only
retiming \cite{kim2026race}. 
RoboPace achieves a $64\%$ overall success rate, compared with $44\%$ for slow
uniform execution, while completing four of five commands in about half its
time.

\section{Related Work}
Imitation-learned policies inherit the temporal sampling of demonstrations
rather than selecting an execution rate for the target robot. This arises in
vision-language-action (VLA) models
\cite{black2024pi0,kim2024openvla,intelligence2025pi05} as well as in earlier,
smaller policies \cite{zhao2023aloha}. Existing work addresses
this mismatch during training or at inference time.

\subsection{Training-time speed adaptation}

One line of work changes the temporal structure of demonstrations before
training. DemoSpeedup down-samples segments according to an entropy-based proxy
for required precision \cite{guo2025demospeedup}, whereas ESPADA uses
semantic judgments to choose the down-sampling schedule
\cite{kim2026espada}. A complementary line makes timing part of the learned
policy: dynamical movement primitives include an explicit temporal constant
\cite{ijspeert2013dmp}; SpeedAug and TempoVLA condition policy execution on
tempo through reinforcement fine-tuning and retimed data, respectively
\cite{nam2026speedaug,jing2026tempovla}; and AutoSpeed learns stage-adaptive
motion speeds \cite{hu2026autospeed}. B-spline Policy instead changes the
action representation to a continuous curve that can be temporally scaled at
execution \cite{han2026bspline}. These methods alter data, training objectives,
or policy representations; adapting learned timing to a new embodiment
generally modifies one of them.

\begin{figure}[!t]
  \centering
  \includegraphics[width=\textwidth]{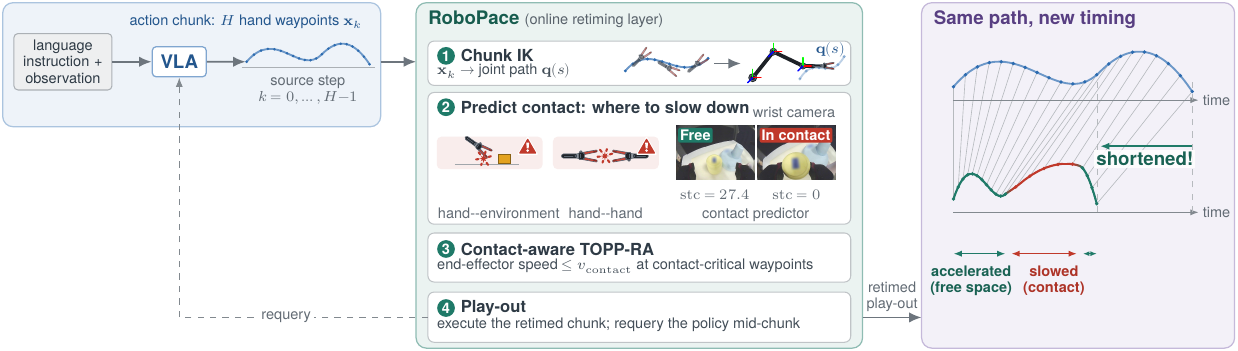}
  \caption{\textbf{RoboPace System Overview:} TOPP-RA retimes each VLA action chunk as
  a joint path, limiting speed at waypoints where contact is predicted. This accelerates free-space motion while
  slowing at contact.}

  \label{fig:overview}
\end{figure}

\subsection{Inference-time speed adaptation}

Time-optimal path parameterization (TOPP) computes the fastest timing of a
fixed geometric path subject to robot constraints. These solvers handle joint velocity,
acceleration, and torque limits via convex
optimization \cite{verscheure2009timeoptimal}, numerical integration
\cite{pham2014topp}, or reachability analysis (TOPP-RA)
\cite{pham2018toppra}. Cartesian constraints, including
end-effector velocity bounds, can likewise be imposed along the path.
TOPP-RA, Cartesian velocity constraints, and proximity-based speed reduction
are established components \cite{marvel2017ssm,haddadin2009safe}. 

RACE \cite{kim2026race} converts predicted action chunks into geometric paths
and applies TOPP-RA under the robot's
kinematic and dynamic constraints; it also uses a test-time search over candidate chunks to
address asynchronous-inference misalignment. While its retiming formulation
uses the robot's kinematic and dynamic constraints, it does not introduce a path-dependent velocity
constraint from predicted contact. SAIL \cite{ranawaka2025sail} adapts speed using criticality cues by switching among preset rates.
SAIL exemplifies adaptive speed modulation, but its rate-selection rule does
not formulate it as a path-parameterization problem subject to
the robot's kinematic and dynamic constraints.
Other inference-time methods address different aspects of action-chunk
execution. Real-time chunking \cite{black2025rtc} and
masked action chunking \cite{wang2026mac} maintain consistency across
asynchronous chunk transitions; PACE \cite{nie2026pace} selects an execution
horizon; and TRACT \cite{liu2026tract} routes chunks across procedural phases.

Prior work has separately addressed two complementary aspects of execution-speed
adaptation: robot-limit-aware time-optimal retiming and criticality-based speed
modulation. To our knowledge,
these elements have not been integrated for online retiming of policy-predicted
action chunks by using predicted contact to define a path-varying velocity
constraint that is jointly enforced with the target robot's kinematic and dynamic constraints.
RoboPace provides this inference-time integration while leaving the policy and
its predicted geometry unchanged, without policy retraining.

\section{Problem Formulation}
\label{sec:problem}
We consider online retiming of policy-predicted action chunks. The goal is to
execute each chunk as fast as permitted by the target robot's kinematic and
dynamic constraints while slowing near predicted contact. At inference time,
the policy outputs $H$ end-effector pose deltas for each arm on the fixed demo-speed source
grid with period $\Delta t_{\mathrm{src}}$. We separate the chunk's geometry from this source
timing. Let $\mathbf{q}(s)\in\mathbb{R}^n$, $s\in[0,1]$, denote the joint-space
spline supplied to the retimer, through the retained waypoints of both arms,
with $s$ uniform over the waypoints. Let $\mathcal{C}\subseteq[0,1]$ denote the contact-critical portion
of the path, comprising waypoints within a threshold of predicted or ongoing
contact. We
seek a monotone time parameterization $s(t)$ by solving
\begin{subequations}
\label{eq:problem}
\begin{align}
  \min_{s(\cdot),\,T}\ \ & T \\
  \text{s.t.}\ \ & s(0)=0,\ \ s(T)=1,\ \ \dot{s}(t)\ge 0, \\
  & |\dot{q}_i(t)|\le v_{\max,i},\ \ |\ddot{q}_i(t)|\le a_{\max,i}
      \quad\ i=1,\dots,n, \label{eq:problem_lim}\\
  & v_b(t)\le v_{\mathrm{contact}}\ \text{for each hand body } b
      \quad\text{if } s(t)\in\mathcal{C}, \label{eq:problem_gate}\\
  & \dot{s}(0)=\dot{s}_0 . \label{eq:problem_entry}
\end{align}
\end{subequations}
Here, $T$ is the chunk duration, $v_{\max,i}$ and $a_{\max,i}$ are the
velocity and acceleration limits of joint $i$, representing the robot's
kinematic and dynamic constraints, $v_b(t)$ is the Cartesian speed of hand
body $b$, $v_{\mathrm{contact}}$ is the contact speed limit, and $\dot{s}_0$ is
the path speed corresponding to the joint velocity commanded at the
transition, allowing consecutive chunks to join without stopping. We solve
\eqref{eq:problem} with TOPP-RA \cite{pham2018toppra} chunk by chunk, with
the entry speed fixed by \eqref{eq:problem_entry} and the exit speed set as in
Sec.~\ref{sec:topp}. The resulting timing is time-optimal per chunk. The contact speed limit is smoothly ramped over
$\rho$ source steps on either side of each contact-critical waypoint
(Sec.~\ref{sec:topp}). The set $\mathcal{C}$ is obtained from geometric
proximity and a learned contact predictor (Sec.~\ref{sec:cues}).

\section{RoboPace}
\subsection{Overview}

\begin{figure}[t]
  \centering
  \includegraphics[width=0.5\linewidth]{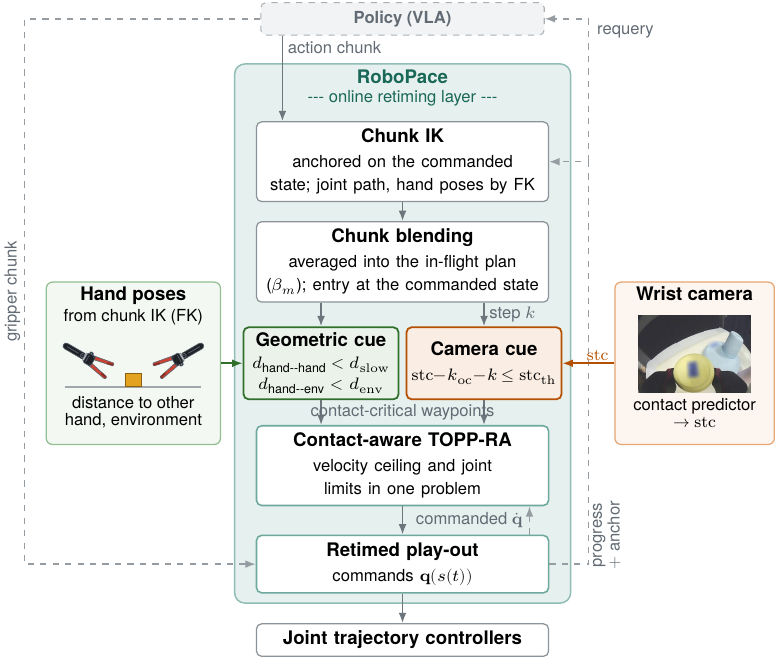}
  \caption{\textbf{RoboPace Pipeline:}
 Geometric and camera cues identify
contact-critical waypoints for TOPP-RA retiming. The play-out feeds its
commanded velocity into the next solve, allowing consecutive action chunks to
join without stopping.}
  \label{fig:pipeline}
\end{figure}

For each predicted action chunk, RoboPace solves the time-optimal retiming
problem in \eqref{eq:problem} online (Fig.~\ref{fig:overview}): each chunk from the deployed policy becomes
the joint path $\mathbf{q}(s)$, cues mark its contact-critical waypoints,
contact-aware TOPP-RA solves $s(t)$, and the play-out commands
$\mathbf{q}(s(t))$ and requeries the policy after a fixed
interval of $N_r$ commanded source steps (Fig.~\ref{fig:pipeline}). The only
learned component of RoboPace is the contact predictor; the rest is
deterministic given fixed parameters and the robot's constraints.

\subsection{Action Chunks to Joint Paths}
\label{sec:ik}

The chunk is converted to joint waypoints through sequential damped-least-squares inverse
kinematics, which preserves the policy's geometry and drops only unreachable
waypoints. Each row, a body-frame pose delta about the hand root, is composed
into an $SE(3)$ target at the tool center point and integrated forward from the
last commanded configuration. Targets are solved within the joint limits, each seeded from the previous
solution, and the path is truncated at the first waypoint whose Cartesian
residual exceeds an IK residual threshold.

The joint paths of the two arms are combined into a single path, with each
waypoint pairing their joint configurations at the same source step. If their
valid IK solutions have different lengths, both are truncated to the shorter
length before being combined. Consecutive waypoints on the combined path that
advance by less than a dwell threshold are then compressed to their first row
while retaining their source indices; this shortens geometric dwells and the
gripper profile sampled over them (Sec.~\ref{sec:playout}). Forward kinematics
provides the world poses of the hand collision bodies at each retained waypoint
for use by the contact cues.

\subsection{Contact Prediction and Gating}
\label{sec:gates}

A waypoint is \emph{contact-critical} when contact is imminent or ongoing: the
retiming holds the hand at or below the contact speed limit
$v_{\mathrm{contact}}$ of \eqref{eq:problem_gate} on contact-critical
waypoints and elsewhere leaves the pace to the robot's kinematic and dynamic
constraints.
We use two complementary cues: geometric proximity and camera-based contact
prediction. The \emph{geometric cue} applies two proximity \emph{conditions},
hand--hand and hand--environment, and flags a waypoint when either distance
falls below its threshold; it requires no sensing but accounts only for modeled
geometry. The \emph{camera cue} predicts the steps to contact $\mathrm{stc}$
from that hand's wrist camera and gripper angle. Here, $\mathrm{stc}$ is the
number of source steps until its next contact and is zero during contact
(Sec.~\ref{sec:cues}). A waypoint is marked contact-critical if either cue is
active.

A prediction refers to the instant $t_{\mathrm{obs}}$ at which its image was
taken, so its countdown is aged before use. Two delays separate that instant
from the waypoint it is applied to: $k_{\mathrm{oc}}$, the number of source steps
the plan commanded between $t_{\mathrm{obs}}$ and the time at which the
chunk is retimed (observation to command), and $k$, the offset of waypoint $k$
from the first row the play-out will command, in source steps (command to
application). The cue thus marks $\mathrm{crit}_{\mathrm{stc}}(k) =
[\,\mathrm{stc} - k_{\mathrm{oc}} - k \le \mathrm{stc}_{\mathrm{th}}\,]$ with
threshold $\mathrm{stc}_{\mathrm{th}}$. Predictions older than a freshness
threshold are dropped, a hand without a fresh prediction contributes no camera
flag, and the predictor is reset whenever the play-out is stopped.

\subsection{Contact-Aware Path Retiming}
\label{sec:topp}

RoboPace converts the Cartesian contact speed limit into a position-dependent
joint velocity ceiling along the path. Defining the ceiling over path position
keeps it tied to the same waypoints when TOPP-RA changes the execution timing.
With $\mathbf{x}_b(s)$ denoting the position of hand body $b$, $\ell'(k)$ is
the hand path-length derivative at waypoint $k$, evaluated from the chord rates
$\lVert\mathbf{x}_b(s_{j+1})-\mathbf{x}_b(s_j)\rVert/(s_{j+1}-s_j)$ of the path
segments $j$ adjacent to $k$ and maximized over them and over the hand
bodies. Requiring every hand body to move at or below $v_{\mathrm{contact}}$
therefore bounds the path speed by $v_{\mathrm{contact}}/\ell'(k)$.

The bound is applied through a gate weight $g(k)$, which rises along a cosine
ramp from $0$ at $\rho$ source steps from the nearest contact-critical waypoint
to $1$ at that waypoint. With $g=g(k)$, the joint velocity ceiling is
\[
v_i(k)=(1-g)\,v_{\max,i}
  +g\min\bigl(v_{\max,i},\;
  v_{\mathrm{contact}}\lvert q_i'(k)\rvert/\ell'(k)\bigr).
\]
When $\ell'(k)\to0$, the hand-body positions are stationary along the path, so
no contact-derived ceiling is imposed there and the joint velocity limits still
apply. Because both arms share one path parameter, a contact constraint on
either hand can slow both arms.

The ceiling replaces $v_{\max,i}$ in \eqref{eq:problem_lim}. We fix the TOPP-RA
boundary speeds so that consecutive chunks join without stopping. At a
transition, the commanded joint velocity $\dot{\mathbf{q}}_{\mathrm{cmd}}$ is
projected onto the new path tangent, $\dot{s}_0 =
\mathbf{q}'(0)^{\top}\dot{\mathbf{q}}_{\mathrm{cmd}}/\lVert\mathbf{q}'(0)\rVert^{2}$,
and clipped into $[0,\,\mu\,\dot{s}_{\max}(0)]$, where $\dot{s}_{\max}(0)$ is
the largest path speed the ceiling admits at $s=0$ and $\mu<1$ a margin. The
exit targets the largest terminal path speed the ceiling admits and is
progressively relaxed until the solve is feasible, so a chunk never decelerates
merely because it ends. If no attempt succeeds, or too few waypoints survive,
the chunk is discarded and the previous plan continues to play.

Away from contact, the pace is set by the robot's velocity and acceleration
constraints along the predicted path, scaled by one factor $\sigma$ as
$\sigma v_{\max}$ and $\sigma^2 a_{\max}$; the demo speed does not constrain
this pace.

\subsection{Online Execution and Chunk Transition}
\label{sec:playout}

The retimed segment is played on a plan clock while progress is tracked in
source steps, so that the policy is requeried after $N_r$ commanded source
steps, however much the retiming stretched them in time. The segment is resampled
uniformly in time and interpolated linearly; a waypoint counts as \emph{commanded} when the plan clock
passes it, according to the source index carried by each retained row, so rows dropped by
the inverse kinematics still advance the count; a wall-clock fallback triggers a
policy requery when a gate hold or a rejected chunk stalls the source-step count.

A new chunk is blended into the in-flight plan because adopting it verbatim
would send its disagreement with the old plan to the arm as a step. With
$\mathbf{q}^{\mathrm{old}}$ the path being played, $\mathbf{q}^{\mathrm{new}}$
the incoming one, $p$ the fractional waypoint the plan clock has reached, and
$\beta_m$ the weight given to the new path at step $m$, the head of the new
path is averaged with the old one read $m$ steps ahead of $p$,
\begin{equation}
\label{eq:blendmix}
\mathbf{q}^{\mathrm{blend}}_m=\beta_m\,\mathbf{q}^{\mathrm{new}}_m
 +(1-\beta_m)\,\mathbf{q}^{\mathrm{old}}(p+m),
\end{equation}
linearly interpolated in the fractional index, the new path standing alone
once the old one ends. The schedule $\beta_m$ runs over blending steps
$m=1,\dots,N_b$ ($\beta_m=1$ beyond $N_b$) and is one of
\begin{equation}
\label{eq:blend}
\begin{aligned}
&\beta_m^{\mathrm{lin}} = \frac{m}{N_b},\qquad
\beta_m^{\mathrm{cos}} = \frac{1-\cos(\pi m/N_b)}{2},\\
&\beta_m^{\mathrm{tanh}} = \frac{\tanh\kappa\bigl(\tfrac{m}{N_b}-\tfrac12\bigr)+\tanh\tfrac{\kappa}{2}}{2\tanh\tfrac{\kappa}{2}}
\end{aligned}
\end{equation}
with shape parameter $\kappa$, as swept in Sec.~\ref{sec:ablation}; a schedule is named by
its weight and window, e.g., Linear-$N3$ for $\beta^{\mathrm{lin}}_m$ with
$N_b=3$. Rows that fall behind the plan clock during pipeline latency are
dropped. The remaining rows are shifted by the displacement of the skipped
prefix---in joint space for the joints and in Cartesian space for the hand
positions, with orientations unchanged---and anchored at the commanded state,
so already commanded motion is not repeated. The new plan clock is then
advanced by the progress the old plan actually made during the solve.

The hand positions that set the ceiling $v_i(k)$ are blended with the same
weights as the joints. The gripper chunk bypasses the inverse kinematics and is
sampled at the source step reached by the arm's plan clock.

\section{Experiments}
We evaluate whether contact-aware retiming enables reliable contact-rich manipulation without the time cost of uniformly slow execution. We compare RoboPace with uniform-rate execution and physical-limit-only retiming across three contact-rich tasks and examine its sensitivity to the contact speed limit and transition schedule. Finally, we qualitatively evaluate RoboPace on a bimanual handover task.

\subsection{Experimental Setup}
\label{sec:setup}

\textbf{Platform and Policy.}
All experiments use $\pi_{0.5}$-based policies
\cite{intelligence2025pi05} on a dual-arm 7-DoF OpenArm v1.0
\cite{enactic2025openarm} with YUBI hands \cite{ohkawa2026yubi}.
Policy inference and retiming run on a workstation with an NVIDIA GeForce
RTX 5070 Ti GPU. The policies are fine-tuned end to end with an action
horizon of $32$, using the two wrist cameras as visual input, from a
checkpoint pretrained on $8{,}434$\,h of YUBI recordings. The test-time instruction exactly matches the instruction used during training.

\textbf{Tasks.}
We evaluate three contact-rich tasks comprising five commands: \task{Cup} places a cup on a plate, \task{Socks} puts a pair of socks into a container, and \task{Pyramid} builds a three-cup pyramid in three separately evaluated phases. The \task{Cup} and \task{Socks} policy is fine-tuned on $8{,}736$ human-collected episodes and the \task{Pyramid} policy on $5{,}288$. We also use \task{Part}, which transfers a finger part from the right to the left hand before placing it in an assembly container, to examine hand-to-hand contact in Sec.~\ref{sec:sps}.

\textbf{Evaluation Protocol.}
A trial succeeds if it reaches the task goal without human intervention
and fails otherwise. For safety, the operator stops a trial before a
collision and records it as a failure; a trial also fails if the task
becomes unrecoverable. We impose no timeout. Task time $t$ is the wall-clock duration from launching the inference process
until it is stopped by the operator, estimated by sampling the process status
at $1$\,Hz.

\begin{figure}[!t]
  \centering
  \includegraphics[width=0.62\linewidth]{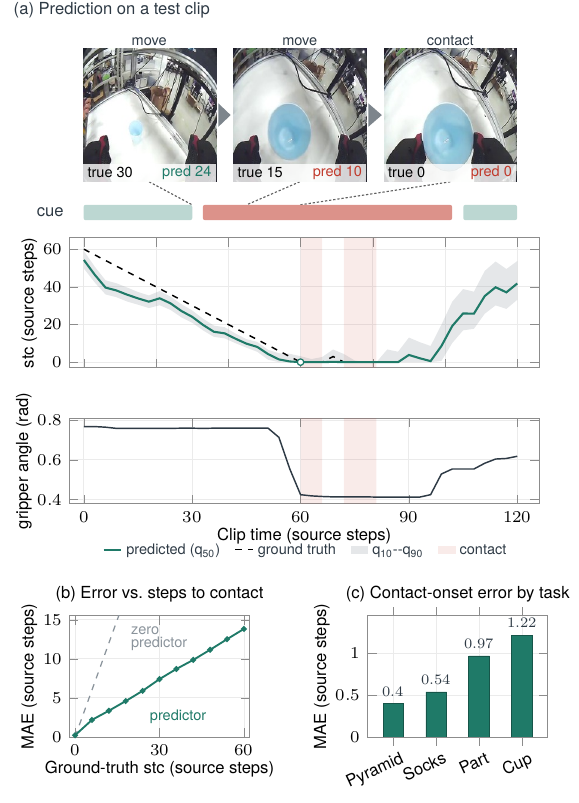}
  \caption{\textbf{Contact Predictor Evaluation:}
  (a) predicted and ground-truth $\mathrm{stc}$ on one \task{Cup} test clip,
  (b) mean absolute error (MAE) versus ground-truth $\mathrm{stc}$, and
  (c) per-task MAE at contact onset
  ($\mathrm{stc}_{\mathrm{gt}}=0$).
  At contact onset, MAE is below $1.3$ source steps for all tasks.}
  \label{fig:cpeval}
\end{figure}

\textbf{Contact Cues.}\label{sec:cues}
For the experiments, we instantiate the contact gate of Sec.~\ref{sec:gates}
with two complementary cues: geometric proximity and learned contact
prediction.

The geometric cue flags a waypoint when hand meshes come within
$d_{\mathrm{slow}}=0.08$\,m of the opposite hand or
$d_{\mathrm{env}}=0.02$\,m of the environment, which we model as the table. 

The learned camera cue predicts $\mathrm{stc}$, the number of source steps
until hand--object contact, from the wrist camera and gripper angle at
$10$\,Hz, using one egocentric model per hand with shared weights; its
threshold is $\mathrm{stc}_{\mathrm{th}}=20$ source steps. The predictor
consists of an EfficientNet-B0 followed by a GRU over the last eight wrist
frames and gripper angles ($4.69$\,M parameters). It is trained on $8{,}601$
recordings drawn from the same human-collected demonstrations used to
fine-tune the VLA policy, with no additional training data, split by recording
into $6{,}021$ for training, $1{,}290$ for validation, and $1{,}290$ held out
for evaluation. The model predicts the $0.1$, $0.5$, and $0.9$ quantiles of
the contact countdown, and we use the median as $\mathrm{stc}$.

Fig.~\ref{fig:cpeval} summarizes contact-predictor performance on the
held-out split. Prediction error decreases as contact approaches. At contact
onset, the first frame of each interval in which the ground-truth countdown
$\mathrm{stc}_{\mathrm{gt}}$ is zero, the median prediction deviates from
$\mathrm{stc}_{\mathrm{gt}}$ by $0.80$ source steps ($0.027$\,s), averaged
over $1{,}540$ onsets.

\begin{figure}[!t]
  \centering
  \includegraphics[width=\textwidth]{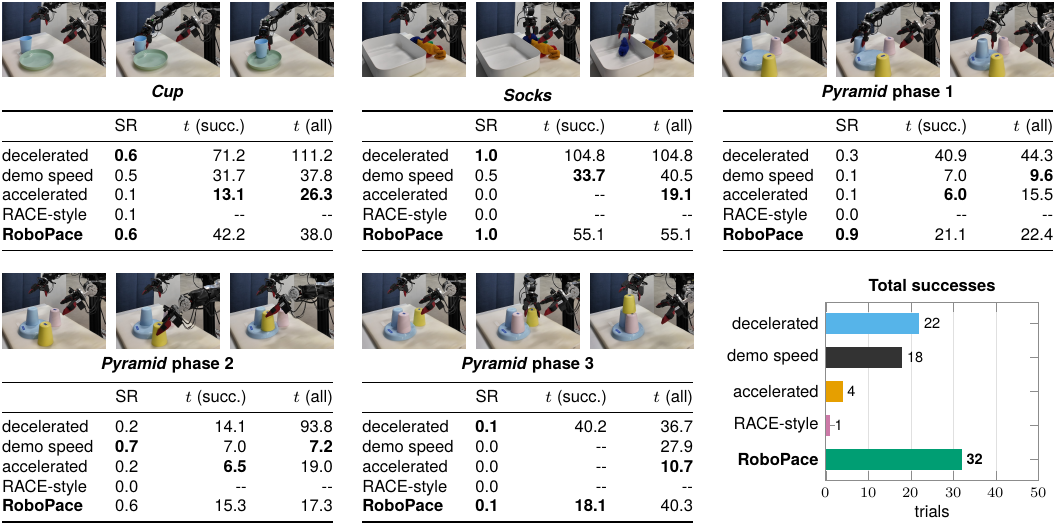}
  \caption{\textbf{Comparison with Baselines:} Accelerated execution and physical-limit-only retiming rarely succeed,
whereas decelerated execution is reliable but slow. RoboPace achieves
$32$ successes in $50$ trials and completes four of five commands in about
half the decelerated execution time.}
  \label{fig:main}
\end{figure}

\textbf{Experimental Configuration.}
Unless otherwise noted, RoboPace uses
$v_{\mathrm{contact}}=0.02$\,m/s, $\sigma=0.9$, $\rho=4$, $H=32$, and the
Linear-$N3$ blend of Sec.~\ref{sec:playout}; for the tanh blend we set
$\kappa=4$. For \task{Socks}, we use
$v_{\mathrm{contact}}=0.04$\,m/s, which remained reliable in pilot runs.
While a hand remains in contact with a held object, the predictor's countdown stays near zero because its training target during contact is zero, so the camera cue keeps that hand contact-limited throughout transport in the main experiments. In the \task{Part} analysis, we enable a \emph{held-hand exclusion}: once a gripper latch marks a hand as held after repeated close commands, that hand's learned contact prediction is excluded from the gate, allowing it to accelerate again after the slowdown hold clears.

\textbf{Implementation Details.}
The layer runs on a $30$\,Hz plan clock matching the source grid, requeries the
policy after $N_r=16$ commanded source steps, resamples each retimed segment to
$64$ points, and triggers a policy requery after $1.6$\,s of wall-clock time if
a gate hold or a rejected chunk stalls the source-step count. The inverse kinematics truncates a chunk where the Cartesian
residual exceeds $0.15$\,m, compresses runs advancing by less than
$10^{-4}$\,rad, and discards a chunk that leaves fewer than four waypoints;
predictions older than $1$\,s are dropped. Each hand is modeled by three collision bodies. The boundary margin is $\mu=0.99$: the entry speed is capped at that fraction of
the admissible path speed and the exit targets the same fraction, relaxed by
factors $1$, $0.75$, $0.5$, $0.25$, and $0$ until the solve succeeds. On this robot, the acceleration
limit is the active motion constraint.

\subsection{Comparison with Baselines}
\label{sec:main}

We compare RoboPace with four baselines that execute the same action chunks (Fig.~\ref{fig:main}): three uniform-rate conditions without retiming---\emph{demo speed}, the $30$\,Hz rollout rate used in training and inherited
directly from the demonstrations; \emph{accelerated} at $50$\,Hz; and \emph{decelerated} at $10$\,Hz---and \emph{RACE-style}, which retimes under the robot's kinematic and dynamic constraints alone. Each condition is evaluated for ten trials per command. Fig.~\ref{fig:main} reports the success rate (SR) and the mean task time over successful trials, $t_{\mathrm{succ}}$, and over all trials $t_{\mathrm{all}}$.

RoboPace retains the reliability of decelerated execution while substantially reducing its time cost: it achieves a higher overall success rate, $32$ versus $22$ successes, with no command below the decelerated rate, and takes only $45$--$59\%$ of its $t_{\mathrm{succ}}$ on four. By comparison, accelerated execution succeeds in only four of fifty trials, while decelerated execution takes two to six times the demo-speed $t_{\mathrm{succ}}$ (excluding \pyrph{3}, where the demo speed never succeeds and $t_{\mathrm{succ}}$ is therefore undefined). Relative to demo speed, RoboPace improves success on four commands. On \task{Socks}, RoboPace takes longer than demo speed. This may partly reflect the task's larger pre- and post-grasp motions, which can amplify temporal mismatch between the policy's predicted gripper closure
and the executed trajectory under chunkwise execution. Increasing the requery
frequency while slowed could reduce this mismatch by incorporating more recent
observations.

\begin{figure}[!t]
  \centering
  \includegraphics[width=\textwidth]{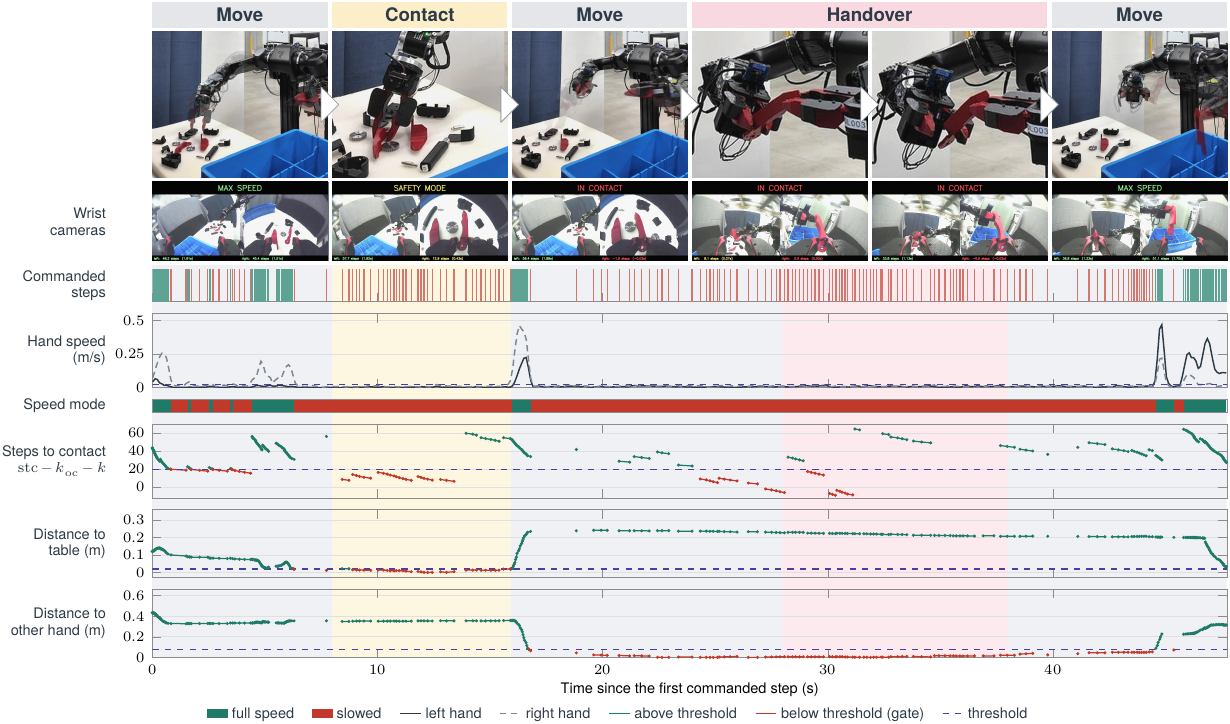}
  \caption{\textbf{Execution Trace of a Bimanual Handover:} One run of \task{Part}, showing that RoboPace slows as the right hand
approaches the part and during hand--hand contact, while accelerating
elsewhere. Red and green indicate gated and ungated motion, respectively.}

  \label{fig:sps}
\end{figure}

\label{sec:prior}%
To isolate the effect of retiming, RACE-style replaces RoboPace's retiming
component with that of RACE \cite{kim2026race} while sharing the policy,
action chunks, and execution stack. It retimes each chunk solely under the
robot's kinematic and dynamic constraints, without RoboPace's contact gate, using RACE's $0.9$
constraint margin and chunk-selection procedure. We do not use RACE's
reached-state retraining. RACE-style succeeds in one of fifty trials
(Fig.~\ref{fig:main}); task times were not recorded for these runs.

Overall, RoboPace retains the reliability of decelerated execution while
substantially reducing its time on four of five commands, whereas faster
uniform execution and physical-limit-only retiming largely fail.

\subsection{Qualitative Analysis of Bimanual Handover}
\label{sec:sps}

We qualitatively evaluate RoboPace on a bimanual handover using one run of
\task{Part} (Fig.~\ref{fig:sps}) to examine how hand--hand contact gating
supports precise contact-rich manipulation within the retiming pipeline. RoboPace slows selectively during the handover,
enabling a successful hand-to-hand transfer, and accelerates again outside the
contact-critical phase. This complements the five quantitative commands, none
of which involves hand--hand contact. 
Unlike the main quantitative experiments, this run uses Tanh-$N9$, requeries with
$N_r=8$, and enables the held-hand exclusion. Because the predictor was not
trained to predict the next contact while a grasp contact is present, the
camera cue provides no slowdown before the final placement;
the geometric cue does not trigger either because it models the
tabletop rather than the raised assembly container. We therefore stop the run before the final placement into the container and use it
to analyze the contact-gating behavior.

Fig.~\ref{fig:sps} quantifies this behavior. The countdown row shows the value tested by the camera cue at each commanded waypoint, $\mathrm{stc}-k_{\mathrm{oc}}-k$ from Sec.~\ref{sec:gates}, with one segment per chunk. Ungated motion proceeds at the demo speed or above, with a median hand speed
of $0.17$\,m/s and peaks near $0.5$\,m/s, whereas gated motion slows to a
median of $0.01$\,m/s. The hand--environment condition is active while the right hand approaches the part, and the camera cue activates as the right hand approaches contact and again during subsequent contact-critical motion. The hand--hand condition limits both hands from $17$ to $45$\,s, through the handover and until the hands separate. Overall, the $299$ source steps take $48$\,s to execute.

\subsection{Retiming Sensitivity}
\label{sec:ablation}

We evaluate RoboPace's sensitivity to the contact speed limit
$v_{\mathrm{contact}}$ and the transition schedule used to blend a new chunk
into the in-flight plan, with the contact gate fixed. We vary both on
\pyrph{2} at $\sigma=0.9$, with five trials per setting. We test the linear,
cosine, and tanh schedules of Sec.~\ref{sec:playout} over $N_b$
steps.

\begin{table}[!t]
\caption{Sensitivity to the contact speed limit and transition schedule.
(a) Five trials per setting on \pyrph{2}: success is more sensitive to the
contact speed limit, with lower limits trading speed for reliability.
(b) Ten trials per command across five commands: Linear-$N3$ (bold) achieves the highest overall
success ($32/50$) among the three lowest-cost schedules selected at
$0.02$\,m/s and is used as the default.}
\label{tab:ablation}
\centering
\footnotesize
\setlength{\tabcolsep}{3pt}
\noindent\rule{\columnwidth}{0.5pt}
\par\vspace{3pt}\noindent\parbox{\columnwidth}{\footnotesize (a) Contact speed limit and
transition schedule: SR ($t_{\mathrm{succ}}$)}
\par\vspace{1pt}
\begin{tabular*}{\columnwidth}{@{}l@{\extracolsep{\fill}}ccc@{}}
 & \multicolumn{3}{c}{$v_{\mathrm{contact}}$ (m/s)} \\
\cmidrule(l){2-4}
schedule & $0.01$ & $0.02$ & $0.04$ \\
\midrule
\textbf{Linear-$N3$} & 0.6 (23.1) & \textbf{0.8 (16.1)} & 0.4 (10.0) \\
Linear-$N9$ & 1.0 (26.3) & 0.2 (13.1) & 0.0 (--) \\
Cosine-$N3$ & 0.8 (25.4) & 0.4 (15.6) & 0.2 (8.0) \\
Cosine-$N9$ & 0.6 (23.8) & 0.8 (22.4) & 0.2 (8.0) \\
Tanh-$N3$   & 1.0 (32.5) & 0.8 (16.9) & 0.2 (9.0) \\
Tanh-$N9$   & 1.0 (27.3) & 0.8 (16.6) & 0.4 (8.5) \\
\end{tabular*}
\par\vspace{2pt}\noindent\rule{\columnwidth}{0.5pt}
\par\vspace{5pt}\noindent\parbox{\columnwidth}{\footnotesize (b) Selected schedules across
five commands: SR ($t_{\mathrm{succ}}$)}
\par\vspace{1pt}
\begin{tabular*}{\columnwidth}{@{\extracolsep{\fill}}lccccc@{}}
 & & & \multicolumn{3}{c}{\task{Pyramid}} \\
\cmidrule(l){4-6}
schedule & \task{Cup} & \task{Socks} & phase 1 & phase 2 & phase 3 \\
\midrule
\textbf{Linear-$N3$} & \textbf{0.6 (42.2)} & \textbf{1.0 (55.1)} & \textbf{0.9 (21.1)} & \textbf{0.6 (15.3)} & \textbf{0.1 (18.1)} \\
Tanh-$N3$   & 0.2 (31.2) & 1.0 (37.5) & 0.8 (33.9) & 0.7 (15.7) & 0.0 (--) \\
Tanh-$N9$   & 0.3 (34.5) & 1.0 (58.6) & 0.6 (19.8) & 0.5 (15.9) & 0.1 (21.1) \\
\bottomrule
\end{tabular*}
\end{table}

Performance is more sensitive to the contact speed limit than to the
transition schedule (Table~\ref{tab:ablation}a). At $0.01$\,m/s, the
schedules achieve three to five successes in five trials, with mean task
times of $23$--$33$\,s; at $0.04$\,m/s, they achieve at most two successes,
with task times of $8$--$10$\,s. At $0.02$\,m/s, four of the six schedules
achieve four successes at $16$--$22$\,s, while the remaining two achieve one
and two. This motivates our choice of $0.02$\,m/s as an intermediate operating
point between the slower, more reliable $0.01$\,m/s setting and the faster,
less successful $0.04$\,m/s setting. At each fixed contact speed limit, changing
the transition schedule changes mean task time by at most ten seconds and
success by two to three trials, with no consistent ranking across the three
limits.

Among the four schedules that achieve four of five successes at
$0.02$\,m/s, we select the three with the lowest \emph{composite cost}:
Linear-$N3$, Tanh-$N3$, and Tanh-$N9$. The cost averages the mean absolute
commanded acceleration, mean absolute commanded jerk, and mean absolute
tracking error, each normalized by the demo-speed condition on the same
command. Across all five commands with ten trials each
(Table~\ref{tab:ablation}b), Linear-$N3$ achieves $32$ successes in $50$
trials, compared with $27$ for Tanh-$N3$ and $25$ for Tanh-$N9$. We therefore
use Linear-$N3$ as the default RoboPace configuration; these trials are
reported as the RoboPace results in Fig.~\ref{fig:main}.

\clearpage
\section{Conclusions and Discussion}
\label{sec:conclusion}
We introduced RoboPace, an inference-time retiming method that addresses the
mismatch between demonstration timing and the execution capabilities of a
target robot. RoboPace preserves the policy's predicted path while converting
predicted contact into a position-dependent velocity ceiling and incorporating
it with the robot's kinematic and dynamic constraints in a single time-optimal
path parameterization. This allows execution timing to adapt online without
policy retraining. Across three contact-rich tasks and five commands on a
dual-arm robot, RoboPace achieved a $64\%$ overall success rate, compared with
$44\%$ for slow uniform execution, while completing four of five commands in
about half the slow-execution time. It also exceeded the success rate at the
demo speed on four of five commands. These results show that contact-aware
retiming can retain the reliability of slow execution while
substantially reducing its time cost.

RoboPace is designed as a system-level retiming layer between policy
prediction and low-level control and is applicable in formulation across policies and
robots, provided the policy supplies multi-step action predictions and the
target robot's kinematic and dynamic constraints are available. Our evaluation
is limited to one robot and one policy family. Future work will evaluate
generalization across robots, policies, tasks, and prediction horizons.

\section*{Acknowledgment}
This paper is based on results obtained from a project, JPNP25015, commissioned
by the New Energy and Industrial Technology Development Organization (NEDO).

\bibliographystyle{plainnat}
\bibliography{refs/references}

\end{document}